\documentclass[letterpaper, 10 pt, conference]{ieeeconf}  % Comment this line out if you need a4paper

\IEEEoverridecommandlockouts                              % This command is only needed if 
\usepackage{graphicx}
\usepackage{subfigure}
\usepackage[ruled,linesnumbered]{algorithm2e}
\usepackage{amssymb}
\usepackage{amsmath}
\usepackage{bbm}
\usepackage{caption}
\usepackage{booktabs}
\usepackage{mathtools}
\usepackage[normalem]{ulem}
\usepackage{array}
\usepackage{url}
\usepackage{multirow}
\usepackage{tabularx}
\usepackage{float}
\usepackage{tikz}
\usepackage{xcolor}
\usepackage{tkz-euclide}
\usepackage{paralist}
\usepackage{calc}
\usepackage{dsfont}
\usepackage{color}
\usepackage{import}
\usepackage[utf8]{inputenc}
\usepackage{flushend}
\usepackage{balance}
\usepackage{cite}
\makeatletter
\let\NAT@parse\undefined
\makeatother
\usepackage{hyperref}

\graphicspath{{img/}}

\title{\LARGE \bf
	Deep Spectral Clustering via Joint Spectral Embedding and Kmeans
}
\author{Wengang~Guo and Wei~Ye*
	\thanks{College of Electronic and Information Engineering, Tongji University, Shanghai, China
		{\tt\small \{guowg, yew\}@tongji.edu.cn}}
	 \thanks{*Corresponding author.}
}

\begin{document}
	\maketitle
	\thispagestyle{empty}
	\pagestyle{empty}

	\begin{abstract}
		Spectral clustering is a popular clustering method. It first maps data into the spectral embedding space and then uses Kmeans to find clusters. However, the two decoupled steps prohibit joint optimization for the optimal solution. In addition, it needs to construct the similarity graph for samples, which suffers from the curse of dimensionality when the data are high-dimensional. To address these two challenges, we introduce \textbf{D}eep \textbf{S}pectral \textbf{C}lustering (\textbf{DSC}), which consists of two main modules: the spectral embedding module and the greedy Kmeans module. The former module learns to efficiently embed raw samples into the spectral embedding space using deep neural networks and power iteration. The latter module improves the cluster structures of Kmeans on the learned spectral embeddings by a greedy optimization strategy, which iteratively reveals the direction of the worst cluster structures and optimizes embeddings in this direction. To jointly optimize spectral embeddings and clustering, we seamlessly integrate the two modules and optimize them in an end-to-end manner. Experimental results on seven real-world datasets demonstrate that DSC achieves state-of-the-art clustering performance.
	\end{abstract}

	\section{Introduction}\label{sec:introduction}
Clustering has been an active research topic over the years, which aims to find a natural grouping of data such that samples within the same cluster are more similar than those from different clusters.
Spectral clustering is one of the most popular clustering methods, which makes few assumptions regarding the shapes of clusters and has a solid mathematical interpretation.
It works by mapping data into the eigenspace of the graph Laplacian matrix and then performing Kmeans clustering \cite{lloyd1982least} on these spectral embeddings. 
However, spectral clustering suffers from two main challenges:
First, constructing the similarity graph for high-dimensional data becomes non-trivial due to the curse of dimensionality.
Second, the spectral embedding step and the Kmeans step are decoupled, which prohibits joint optimization for achieving better solutions.

\begin{figure}[tbp]
	\centering
	\hspace*{\fill}
	\subfigure[AE, (86.5, 70.5)]{\includegraphics[width=0.315\columnwidth]{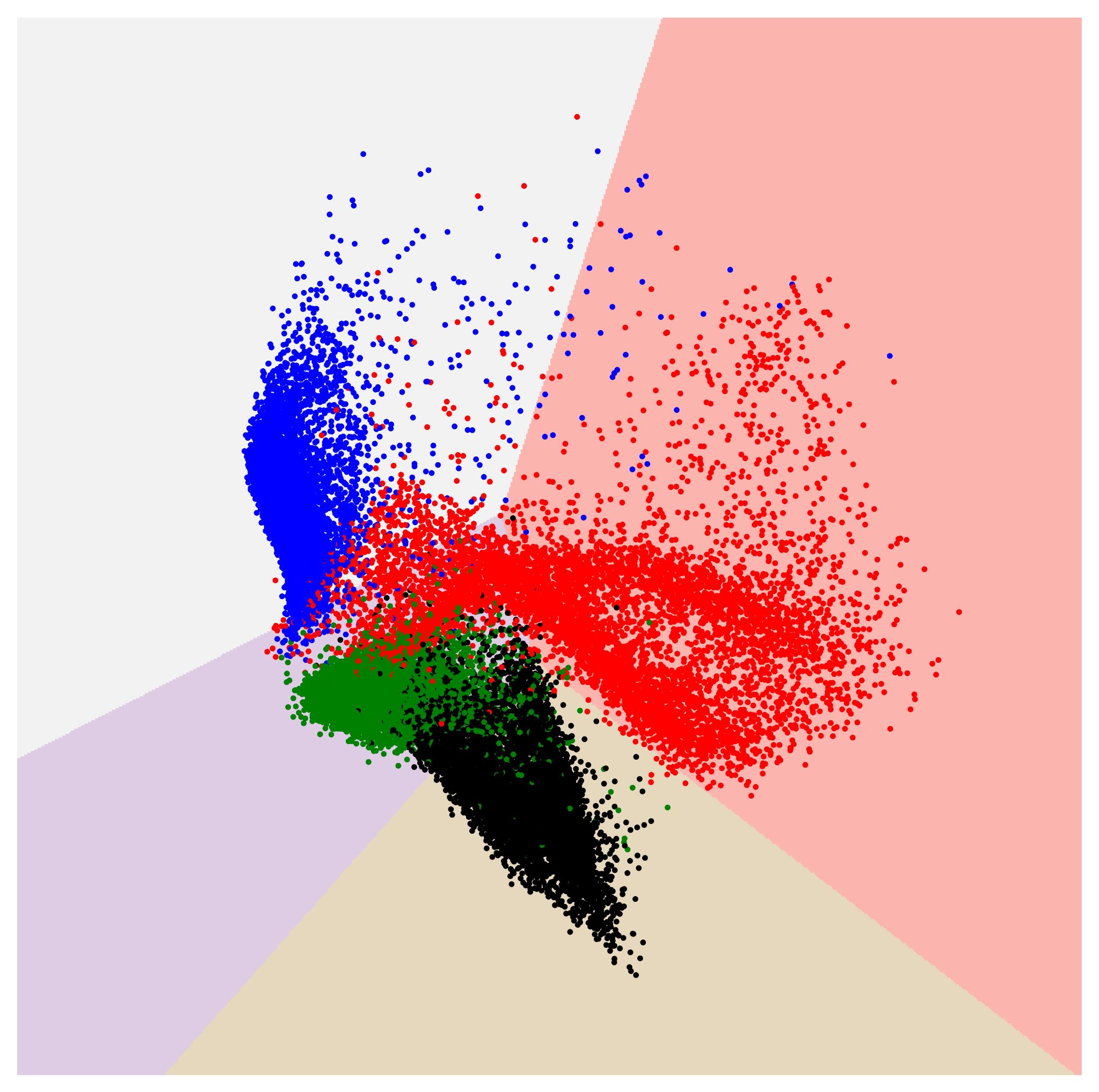}\label{fig_2D_a}}
	\hfill
	\hfill
	\subfigure[Ite 0, (86.7, 71.0)]{\includegraphics[width=0.315\columnwidth]{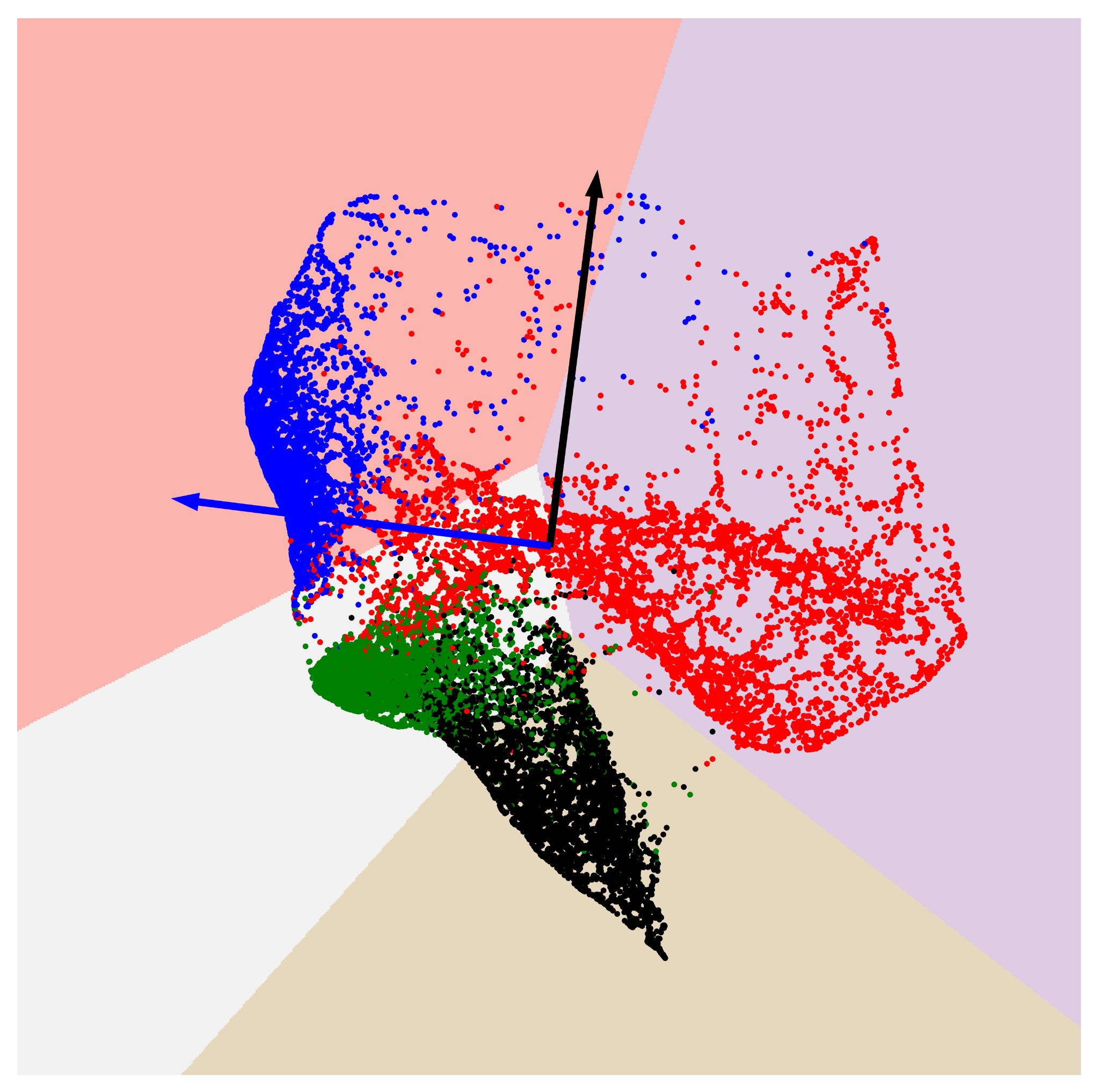}\label{fig_2D_b}}
	\hfill
	\hfill
	\subfigure[Ite 15, (95.1, 84.9)]{\includegraphics[width=0.315\columnwidth]{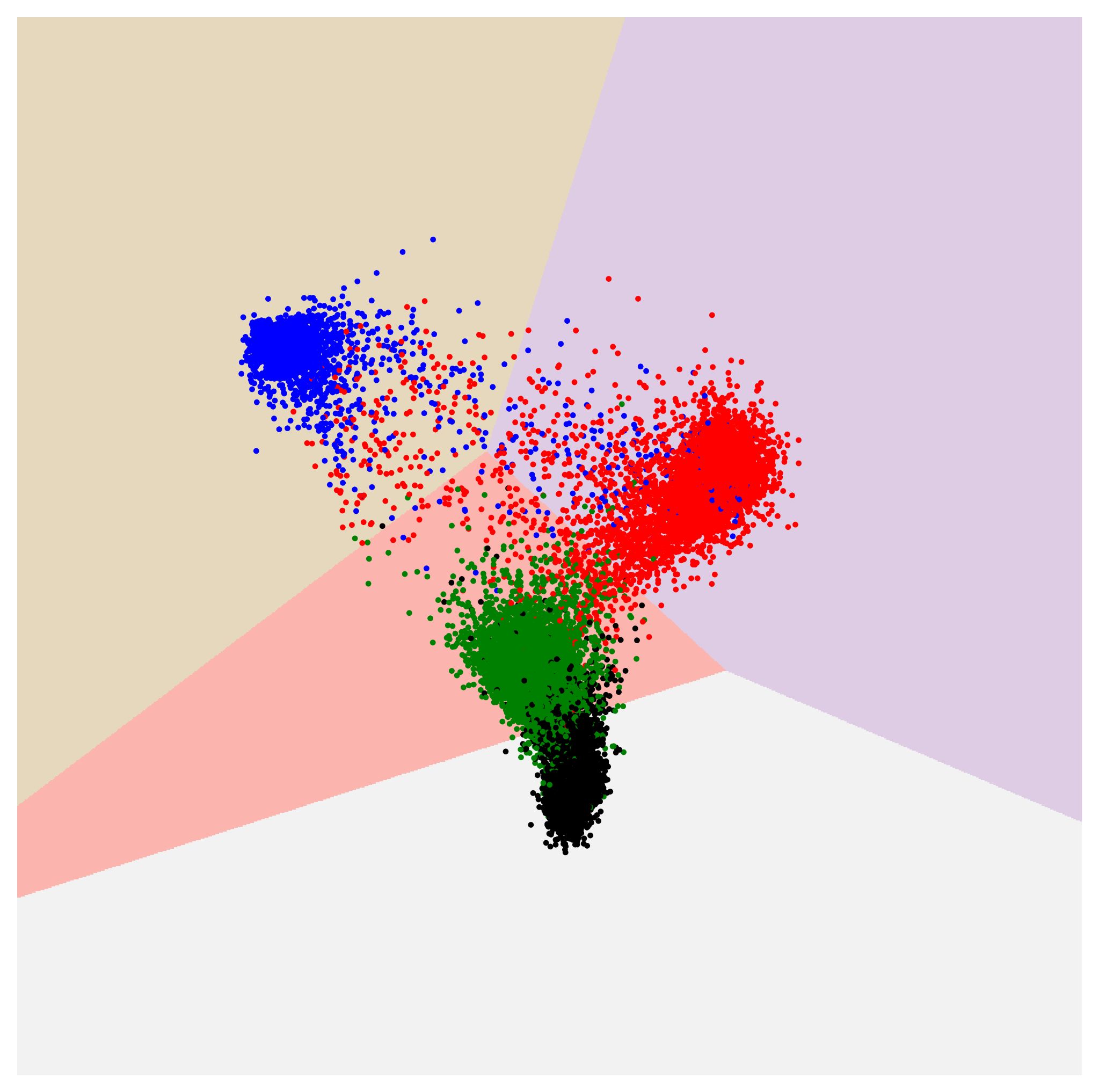}\label{fig_2D_c}}
	\hspace*{\fill}
	\caption{
		Embedding visualization at different training iterations (Ite) of DSC on a subset of the FASHION dataset \cite{xiao2017fashion}. 
		%		(a) shows the initial pretrained autoencoder (AE) embeddings; (b) shows the initial spectral embeddings of our method DSC; (c) shows the spectral embeddings of DSC after one training iteration; (d) shows the final learned spectral embeddings of DSC. 
        We set the number of neurons in the embedding layer of autoencoder (AE) to two for direct 2D visualization.
		The color of samples denotes the ground-truth clusters whereas the background color denotes the Kmeans clustering results. 
		%		The black arrow in (b) shows the direction of the largest eigenvector of the within-class scatter matrix of Kmeans while the blue arrow shows the direction of the smallest eigenvector.
		Numbers in parentheses denote the values of clustering evaluation metrics ACC\% and NMI\%, respectively.
	}
	\label{fig_2D}
\end{figure}

To address the first challenge, the simplest way is to conduct dimensionality reduction before the similarity graph construction.
The early works \cite{wang2015spectral} employ shallow methods for dimensionality reduction, such as Principal Component Analysis (PCA) \cite{pearson1901liii} and Non-negative Matrix Factorization (NMF) \cite{lee1999learning}.
However, these shallow methods cannot adequately capture the complex and nonlinear structures hidden in high-dimensional data due to their limited representational ability.
% Recent efforts have started to use powerful deep neural networks.
% For example, Graphencoder \cite{tian2014learning} exploits a deep autoencoder to map the graph Laplacian matrix into the spectral embedding space. 
% Furthermore, SpectralNet \cite{shaham2018spectralnet} directly maps raw samples into the spectral embedding space. 
% However, these methods require a predefined meaningful similarity matrix for the subsequent spectral embedding learning, but constructing such a matrix is inherently challenging for high-dimensional data.
To solve the second challenge, existing methods \cite{yang2016unified,pang2018spectral} typically discretize the continuous spectral embedding matrix into the binary cluster assignment matrix using a linear transformation.  
However, the discreteness error could be inevitably large, which results in suboptimal clustering performance.

Recently, some literature has started to employ powerful deep neural networks for improving spectral clustering.
Graphencoder \cite{tian2014learning} exploits a deep autoencoder to map the graph Laplacian matrix into the spectral embedding space. 
Furthermore, SpectralNet \cite{shaham2018spectralnet} directly maps raw samples into the spectral embedding space. 
However, these methods require a predefined meaningful similarity matrix for subsequent spectral embedding learning, but constructing such a similarity matrix is inherently challenging for high-dimensional data.
In additions, these methods simply conduct posthoc Kmeans clustering on their learned spectral embeddings for final clustering results, implicitly assuming these embeddings follow isotropic Gaussian structures. 
This assumption is not always valid or reasonable for many real-world data.
% often does not hold for many real-world data, leading to suboptimal clustering performance.
 % is not always valid or reasonable for different types of data structures.

In this paper, we extend spectral clustering to a deep version called DSC to solve the two challenges mentioned above. 
DSC consists of two main modules: the spectral embedding module and the greedy Kmeans module. 
The former module efficiently embeds raw samples into a discriminative low-dimensional spectral embedding space using a deep autoencoder, which is similar to the spectral embedding step in spectral clustering but has a much lower time complexity. 
The high efficiency is attributed to replacing the labor-intensive eigendecomposition with the lightweight power iteration method.
The latter greedy Kmeans module aims to make spectral embeddings Kmeans-friendly.
The motivation is that while the spectral embeddings  derived by graph Laplacian matrix are discriminative, they may not be optimally suited for Kmeans clustering, as the Kmeans objective is agnostic during their generation.
For example, the spectral embeddings generated in Fig. \ref{fig_2D_b} (viewing autoencoder (AE) embeddings in Fig. \ref{fig_2D_a} as inputs) exhibit clear cluster structures, but they do not follow the isotropic Gaussian structures that Kmeans prefers.
To address this issue, we propose fusing the Kmeans objective into the generation process of the spectral embeddings and propose a novel optimization strategy.
Fig. \ref{fig_2D_c} displays the spectral embedding after fusing Kmeans prior, showcasing cluster structures that are more aligned with Kmeans.
To jointly optimize spectral embeddings and clustering, we seamlessly unify these two modules into a joint loss function and optimize this loss in an end-to-end manner.

Our main contributions can be summarized as:
\begin{itemize}
\item We propose a deep spectral clustering method DSC that jointly optimizes spectral embeddings and Kmeans clustering by minimizing a novel joint loss. 
\item To the best of our knowledge, this is the first work of deep joint spectral clustering. 
\item Experiments on 7 real-world datasets demonstrate that DSC achieves state-of-the-art clustering performance.
\end{itemize}

	\section{Related Work}
In this section, we briefly review related works on spectral clustering, power-iteration-based clustering, deep spectral clustering, and deep clustering.

\textbf{Spectral Clustering} formulates the clustering task as a graph cut problem known as Min-Cut, drawing on the correlation between the eigenvalues of the graph Laplacian matrix and the connectivity of a graph.
However, directly solving the Min-Cut problem yields degenerate clustering where a single outlier vertex forms a cluster.
To promote balanced clusters, SC-Ncut~\cite{shi2000normalized} and SC-NJW \cite{ng2002spectral} respectively introduce normalized cut, which have gained much popularity and promote further extensions. 

\textbf{Power-iteration-based clustering} aims to reduce the computational cost of the eigendecomposition in spectral clustering. 
PIC \cite{lin2010power} first proposes exploiting truncated power iteration to compute spectral embeddings.
\cite{boutsidis2015spectral} provides a rigorous theoretical justification for power-iteration-based clustering methods.
To reduce the redundancy of embeddings generated by power iteration, \cite{huang2015diverse} introduces a procedure to orthogonalize these embeddings, and FUSE \cite{ye2016fuse} utilizes Independent Component Analysis (ICA)  \cite{learned2003ica} to make these embeddings pair-wise statistically independent. 
However, these methods still suffer from the high-dimensional data due to the shallow nature of their models.
In contrast, we exploit the high representational power of deep neural networks to cope with the high-dimensional data. 

\textbf{Deep Spectral Clustering} \cite{tian2014learning,shaham2018spectralnet,householder1958unitary} combines spectral clustering with deep neural networks to improve spectral clustering. 
In the viewpoint of matrix reconstruction, Graphencoder \cite{tian2014learning} shows that both spectral clustering and autoencoder aim for the optimal low-rank reconstruction of the input affinity matrix. It thus proposes optimizing the reconstruction loss of the autoencoder instead of eigendecomposition. 
SpectralNet \cite{shaham2018spectralnet} learns to map raw samples into the eigenspace of the graph Laplacian matrix by deep neural networks. 
To prevent trivial solutions, it exploits QR decomposition \cite{householder1958unitary} to ensure embedding orthogonality.
Unlike these methods, we uniquely employ power iteration for spectral embedding learning.

\textbf{Deep Clustering} \cite{DEC_xie2016unsupervised,yang2016joint,ghasedi2017deep,zhang2021learning,cai2022efficient,kwon2023image,zhang2023clusterllm,miklautz2022deep,leiber2022dipencoder,leiber2021dip,miklautz2021details} has recently attracted significant attention. 
For example, DEC \cite{DEC_xie2016unsupervised} pretrains an autoencoder with the reconstruction loss and performs Kmeans to obtain the soft cluster assignment of each sample. Then, it derives an auxiliary target distribution from the current soft cluster assignments, which emphasizes samples assigned with high confidence. Finally, it updates parameters by minimizing the Kullback–Leibler divergence between the soft cluster assignment and the auxiliary target distribution. The key idea behind DEC is to refine the clusters by learning from high-confidence assignments. 
JULE \cite{yang2016joint} introduces a recurrent framework for joint representation learning and clustering, where clustering is conducted in the forward pass and representation learning is conducted in the backward pass.
DEPICT \cite{ghasedi2017deep} consists of an autoencoder for learning the embedding space and a multinomial logistic regression layer functioning as a discriminative clustering model. DEPICT defines a clustering loss function using relative entropy minimization, regularized by a prior for the frequency of cluster assignments. 
As for deep subspace clustering, SENet \cite{zhang2021learning} and EDESC \cite{cai2022efficient} employ the neural network to learn a self-expressive representation. 
In very recent, some literature \cite{kwon2023image,zhang2023clusterllm} has explored Large Language Models (LLMs)  \cite{achiam2023gpt} for clustering.
Due to space limitations, we only review some classical methods here and interested readers can refer to \cite{zhou2022comprehensive}  for a detailed review of deep clustering.

	\section{Preliminaries}

\begin{figure*}[t]
	\centering
	\centerline{\includegraphics[width=0.95\textwidth]{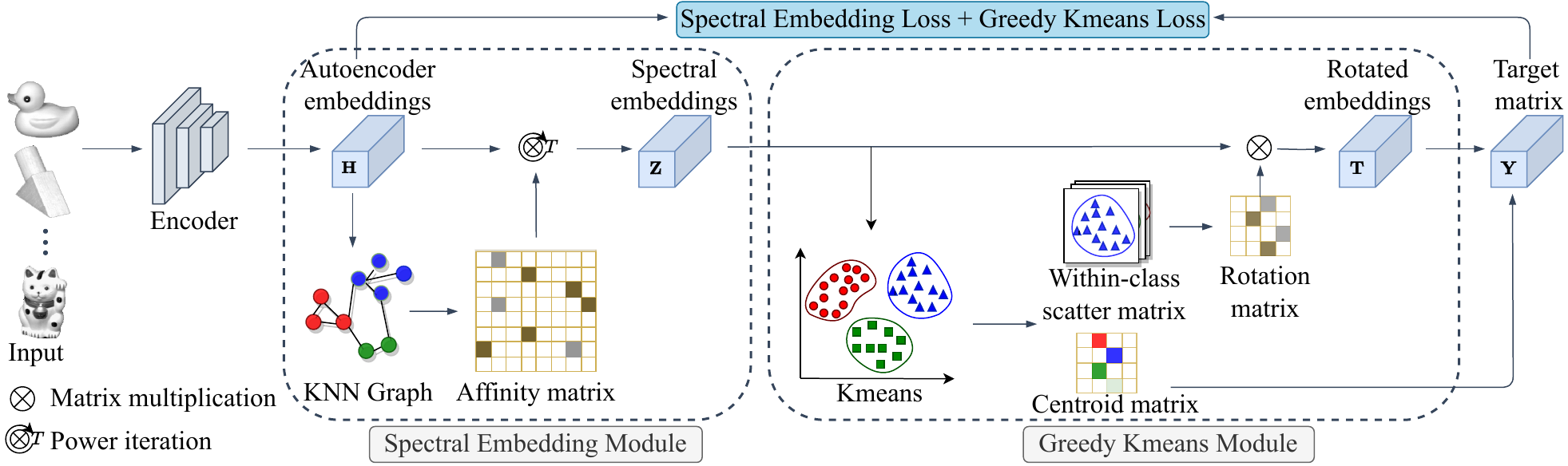}}
	\caption{
		The overall architecture of the proposed DSC.
	}
	\label{fig_framework}
\end{figure*}

\subsection{Spectral Clustering}
Let's consider the problem of grouping $N$ samples $\mathbf{X}=[\mathbf{x}_1,\cdots,\mathbf{x}_N]$ into $K$ distinct clusters.
The classical spectral clustering method SC-Ncut \cite{shi2000normalized} converts the clustering task as a graph cut problem.
Let $\mathbf{A}\in \mathbb{R}^{N\times N}$ denote the affinity matrix where element $\mathbf{A}_{ij}$ represents the similarity between samples $\mathbf{x}_i$ and $\mathbf{x}_j$, $\mathbf{D}=\operatorname{diag}(\mathbf{A}\mathbf{1})$ denote the degree matrix where $\mathbf{1}$ is the vector with all ones,  $\mathbf{W}=\mathbf{D}^{-1}\mathbf{A}$  denote the normalized affinity matrix, and $\mathbf{L}=\mathbf{I}-\mathbf{W}$ denote the normalized graph Laplacian matrix where $\mathbf{I}$ is the identity matrix.
SC-Ncut optimizes the relaxed normalized cut objective:
\begin{equation}
	\label{eqn:relaxedNcut}
\min_{\mathbf{C}\in \mathbb{R}^{N\times K}}\operatorname{Tr}(\mathbf{C}^\intercal \mathbf{L}\mathbf{C}) 	\quad \text{s.t.} ~ \mathbf{C}^\intercal\mathbf{C}=\mathbf{I}_K
\end{equation}
where $\mathbf{C}$ is the spectral embedding matrix (a relaxation of the discrete cluster assignment matrix), $\operatorname{Tr}(\mathbf{C}^\intercal \mathbf{L}\mathbf{C}) $ is the cost of graph cut, and the constraint item encourages $K$-way partition.
SC-Ncut performs eigendecomposition to solve $\mathbf{C}$ that turns out to be the top $K$ minimum eigenvectors of $\mathbf{L}$ (which are also the top $K$ maximum eigenvectors of $\mathbf{W}$).
Finally, SC-Ncut applies Kmeans to cluster the rows of $\mathbf{C}$ and the resulting cluster assignments are assigned back to the original samples.

\subsection{Pretrained Autoencoder Embeddings}
\label{sec:step1}
Autoencoder is a deep neural network that can unsupervisedly learn meaningful representations of data. It commonly consists of a trainable encoder $f(\cdot)$ that maps samples $\mathbf{X}$ into a low-dimensional embedding space $\mathbf{H}$ and a mirrored decoder $g(\cdot)$ that reconstructs the original samples $\mathbf{X}$ from the embedding space $\mathbf{H}$.
Autoencoder is trained by minimizing the reconstruction loss:
\begin{equation}
	\label{eqn:ae}
	\begin{aligned}
		\mathcal{L}_{\text {recon}} &= \left\|\mathbf{X}-g(f(\mathbf{X}))\right\|^2_F
	\end{aligned}
\end{equation}

	\section{Deep Spectral Clustering}
In this section, we elaborate on the proposed deep spectral clustering (DSC) method. 
As shown in Fig. \ref{fig_framework}, DSC has two main modules: the spectral embedding module and the greedy Kmeans module.

\subsection{Spectral Embedding Module}
\label{sec:step2}
Given data $\mathbf{X}=[\mathbf{x}_1,\cdots,\mathbf{x}_N]$, we first pretrain a deep autoencoder to generate the $D$-dimensional embeddings $\mathbf{H}=[\mathbf{h}_1,\cdots,\mathbf{h}_N]\in \mathbb{R}^{N\times D}$.
We then compute the  self-tuning affinity matrix \cite{zelnik2004self}:
\begin{equation}
	\mathbf{A}_{ij}=\exp \Bigl( { - \frac{{{{\left\| {{{\mathbf{h}}_i} - {{\mathbf{h}}_j}} \right\|}^2}}}{{{\sigma _i}{\sigma _j}}}} \Bigr)
\end{equation}
where $\mathbf{h}_i$ is the autoencoder embedding of sample $\mathbf{x}_i$, ${\sigma _i} = \left\| {\mathbf{h}_i - \mathbf{h}_i^M} \right\|$ is a scaling parameter that reflects local density information of $\mathbf{x}_i$, and $\mathbf{h}_i^M$ denotes the $M$-nearest neighbor of $\mathbf{h}_i$.
We also compute the normalized affinity matrix $\mathbf{W}=\mathbf{D}^{-1}\mathbf{A}$.

Computing spectral embedding in Equation \eqref{eqn:relaxedNcut} via eigendecomposition is cost prohibitive when the number of samples $N$ is very large, due to its troublesome time complexity $\mathcal{O}(N^{3})$.
Instead, we propose using power iteration to compute pseudo-eigenvectors, which has a much lower cost, requiring only a few matrix multiplications.
% The resulting pseudo-eigenvectors are linear combinations of the dominant eigenvectors of $\mathbf{W}$ and contain rich cluster separation information.
Specifically, let $\mathbf{H}^{(0)}=\mathbf{H}$, we repeatedly perform the below update rule:
\begin{equation}
	\label{equ_pi_H}
	\mathbf{H}^{(t)}=\mathbf{W}\mathbf{H}^{(t-1)}, \quad t=1,2,\cdots,T
\end{equation}
The final output $\mathbf{H}^{(T)}$ is termed spectral embeddings in this paper, which are linear combinations of the dominant eigenvectors of $\mathbf{W}$ and contain rich cluster separation information. 
The multiplication $\mathbf{W}\mathbf{H}$ can be regarded as $D$ independent matrix-vector multiplication $\mathbf{W}\mathbf{H}_d,d=1,\cdots,D$, where the vector $\mathbf{H}_d\in \mathbb{R}^{N\times 1}$ denotes the $d$-th column of $\mathbf{H}$.
Let $[\lambda_{1}, \cdots,\lambda_{N}]$ and $\mathbf{U}=[\mathbf{u}_{1}, \cdots, \mathbf{u}_{N}]$ denote the eigenvalues and eigenvectors of $\mathbf{W}$ respectively, we can rewrite the $d$-th column of spectral embedding $\mathbf{H}^{(T)}_d$ as:
\begin{equation}
	\label{eqn:vt}
	\begin{aligned}
		\mathbf{H}^{(T)}_d &=\mathbf{W}\mathbf{H}^{(T-1)}_d =\mathbf{W}^{2}\mathbf{H}^{(T-2)}_d =\cdots=\mathbf{W}^{T}\mathbf{H}^{(0)}_d  \\ &=c_{1} \mathbf{W}^{T}\mathbf{u}_{1} +c_{2} \mathbf{W}^{T}\mathbf{u}_{2}+\cdots+c_{N} \mathbf{W}^{T}\mathbf{u}_{N}\\ &=c_{1} \lambda_{1}^{T} \mathbf{\mathbf{u}}_{1}+c_{2} \lambda_{2}^{T} \mathbf{\mathbf{u}}_{2}+\cdots+c_{N} \lambda_{N}^{T} \mathbf{\mathbf{u}}_{N} 
	\end{aligned}
\end{equation}
where $\mathbf{H}^{(0)}_d=\sum_{i=1}^{N}c_{i} \mathbf{u}_{i}$ represents the decomposition of $\mathbf{H}^{(0)}_d$ using the basis $\mathbf{U}$ and $c_{i}$ is the weight coefficient. 
As there exists an eigengap between the $K$-th and $(K+1)$-th eigenvalues (a basic assumption of spectral clustering \cite{shi2000normalized}), $\mathbf{H}^{(T)}_d$ will be a linear combination of the top $K$ maximum eigenvectors of $\mathbf{W}$  after several iterations, and the ablation experiment in Table \ref{tab_ablation} indicate they are highly discriminative.

Choosing the iteration number $T$ is important for clustering, which trades off discrimination power with computational cost.
An excessively large $T$ makes final spectral embeddings converge to the largest eigenvector of $\mathbf{W}$, which is a uniform vector and thus useless for clustering.
To avoid this trivial solution, we introduce the acceleration metric $a^{(t)}=\|\mathbf{v}^{(t)}-\mathbf{v}^{(t-1)}\|_{\infty}$ to dynamically determine the appropriate $T$, where $\mathbf{v}^{(t)}=(\mathbf{H}^{(t)}-\mathbf{H}^{(t-1)})^\intercal\mathbf{1}$.
We set $T$ to be the smallest value of $t$ such that $a^{(t)} \leqslant \hat{a}$, where $\hat{a}$ is a predefined threshold.
For conciseness, the final spectral embeddings $\mathbf{H}^{(T)}$ are hereafter denoted as $\mathbf{Z}$. 

% The orthogonality of spectral embeddings is crucial for detecting multiple clusters ($K>2$). After obtaining the spectral embeddings $\mathbf{H}^{(T)}$, we find an orthogonal matrix $\mathbf{Z}$ that most accurately approximates $\mathbf{H}^{(T)}$ by solving the following optimization problem:
% \begin{equation}
% 	\min_{\mathbf{Z}^\intercal\mathbf{Z}=\mathbf{I}_D}\|\mathbf{H}^{(T)}-\mathbf{Z}\|_F^2
% \end{equation}
% The above equation represents a specific version of the orthogonal Procrustes problem \cite{schonemann1966generalized}. This problem can be solved through the singular value decomposition (SVD) of $\mathbf{H}^{(T)}=\mathbf{U}\mathbf{\Sigma}\mathbf{V}^{\intercal}$, where $\mathbf{U}$ and $\mathbf{V}$ are respectively the left-singular and right-singular vector matrices, and $\mathbf{\Sigma}$ contains the singular values. The solution for $\mathbf{Z}$ is then given by $\mathbf{Z}=\mathbf{U}\mathbf{V}^{\intercal}$.
% This orthogonalization procedure helps to reduce redundancy among the columns of $\mathbf{H}^{(T)}$, thereby improving the discriminative power for clustering.

Generalizing spectral embeddings to previously-unseen samples is non-trivial for traditional spectral clustering methods.
To address this issue, we propose training the autoencoder to directly predict spectral embeddings, which can be formulated as a regression problem:
\begin{equation}
	\label{eqn:E_H}	
	\begin{aligned}
		\mathcal{L}_{\text {spectral}} &= \left\|f(\mathbf{X}) - \mathbf{Z} \right\|^2_F
	\end{aligned}
\end{equation}
In this way, the encoder learns a deep mapping that embeds raw data into their spectral embedding space.
Once trained, the deep mapping can be applied to out-of-sample data without needing to compute and eigendecompose their graph Laplacian matrix as shown by the experimental results in Table \ref{tab_oose}.

\subsection{Greedy Kmeans Module}
After obtaining the spectral embeddings, traditional spectral clustering employs Kmeans to yield the final clustering results.
This may result in underperformed clustering results, as the Kmeans objective is independent to the generation of these embeddings.
We propose fusing the Kmeans objective into the generation process of the spectral embeddings, thereby obtaining Kmeans-friendly embeddings and potentially resulting in better clustering performance.
Specifically, we employ Kmeans to find a partition of the spectral embeddings $\mathbf{Z}=\left[\mathbf{z}_{1} ; \cdots ; \mathbf{z}_{N}\right]$. The Kmeans objective is to minimize:
\begin{equation}
	\label{eqn:kmeans}
	\mathcal{L}_{\text{Kmeans}}  = \sum_{k = 1}^K \sum_{\mathbf{z} \in \mathcal{C}_k} \left\| \mathbf{z} - \boldsymbol{\mu}_k \right\|^2
\end{equation}
where $\mathcal{C}_k$ denotes the set of samples assigned to the $k$-th cluster, $\boldsymbol{\mu}_k=\frac{1}{\left| \mathcal{C}_k\right| }\sum_{\mathbf{z} \in \mathcal{C}_k}\mathbf{z}$ denotes the centroid of $\mathcal{C}_k$.

To generate Kmeans-friendly embeddings, a direct idea is to minimize Objective \eqref{eqn:kmeans} by pulling samples closer to their respective cluster centroids.
% This idea is problematic, as a global optimal solution is $\mathbf{z}\equiv \mathbf{0}$ and the optimal objective value $	\mathcal{L}_{\text{Kmeans}}=0$ can always be achieved. 
To this end, we propose a greedy optimization strategy: pulling samples towards their cluster centroids only in the direction that exhibits the worst cluster structures.
% .rather than in all directions.
This greedy strategy is easier to optimize than naively pulling samples in all directions.
% and effectively prevents the trivial solution by preserving most cluster structures.
To find the direction of the worst cluster structure in the spectral embedding space $\mathbf{Z}$, we rotate $\mathbf{Z}$ using an orthogonal matrix $\mathbf{V}=[\mathbf{v}_1,\cdots,\mathbf{v}_D]^\intercal\in \mathbb{R}^{D\times D}$  ($\mathbf{V}\mathbf{V}^\intercal =\mathbf{I}$).
We select the linear rotation transformation for two reasons: 
(1) the linearity is sufficient for this task, as the deep autoencoder has already learned the non-linear relationships.
(2) the orthogonal matrix maintains the Euclidean distances between embeddings and thus preserves all cluster structures in $\mathbf{Z}$.
We use the Kmeans objective as a prior---the smaller the Kmeans objective value, the better the cluster structures---to help solve for $\mathbf{V}$.
The Kmeans objective value along the direction of $\mathbf{v}_d, d=1,\cdots,D$ is:
\begin{equation}
	\begin{split}
		\mathcal{L}_{\text{Kmeans}}^{d\text{-th}}&={ \sum_{k = 1}^K \sum_{\mathbf{z} \in \mathcal{C}_i} 	\left\| \mathbf{z}\mathbf{v}_d^\intercal - \boldsymbol{\mu}_k\mathbf{v}_d^\intercal \right\|^2}\\
		&=\mathbf{v}_d \underbrace{ \Bigl(\sum_{k = 1}^K \sum_{\mathbf{z} \in \mathcal{C}_k}	 \left(\mathbf{z} - \boldsymbol{\mu}_k\right)^\intercal\left(\mathbf{z}- \boldsymbol{\mu}_k\right)\Bigr)}_{:=\mathbf{S}}\mathbf{v}_d^\intercal
	\end{split}
	\label{eqn_subspace}
\end{equation}
where $\mathbf{S}$ is the within-class scatter matrix of Kmeans.
The Kmeans Objective \eqref{eqn:kmeans} w.r.t $\mathbf{V}$ thus can be rewritten as:
\begin{equation}
	\mathcal{L}_{\text{Kmeans}}
	=\sum_{d = 1}^D\mathbf{v}_d \mathbf{S}\mathbf{v}_d^\intercal
	=\mbox{Trace}\left(\mathbf{V}\mathbf{S}\mathbf{V}^\intercal\right)
	\label{eqn_km_total}
\end{equation}

The solution of $\mathbf{V}$ contains the eigenvectors of $\mathbf{S}$, and the eigenvalues (the Kmeans objective value) indicate the quality of cluster structures in the corresponding eigenvectors. 
The smaller the eigenvalue, the better the cluster structures along its eigenvector direction.
The solution can always be found as $\mathbf{S}$ is symmetric and thus orthogonally diagonalizable, and the computational cost is negligible as the size of $\mathbf{S}$ is much smaller than the raw data, i.e., $D\ll N$.
We sort these eigenvectors in $\mathbf{V}$ in ascending order w.r.t. their eigenvalues. 
Thus, the direction of the last eigenvector $\mathbf{v}_D$ whose eigenvalue is the largest, has the worst quality of the cluster structure. 
Fig. \ref{fig_2D_b} displays an example, where the black arrow denotes $\mathbf{v}_D$.

\begin{algorithm}[t]
	\caption{DSC Algorithm}
	\label{algo}	
	Pretrain an autoencoder by minimizing Loss \eqref{eqn:ae}\;
	\While{not converged}{
		Compute spectral embeddings $f(\mathbf{X})$\;
		Perform Kmeans to obtain clustering results\;
		Update the encoder by minimizing the joint Loss \eqref{eq_loss_all} using the mini-batch SGD optimization\;
	}
\end{algorithm}

After obtaining the largest eigenvector $\mathbf{v}_D$, we will pull samples  towards their cluster centroids in the direction of $\mathbf{v}_D$.
Specifically, we construct a target matrix $\mathbf{Y}=\mathbf{Z}\mathbf{V}\in \mathbb{R}^{N\times D}$ and then replace the last column  of $\mathbf{Y}$ with a pulling target vector $\mathbf{y}=[y_1,\cdots,y_N]\in \mathbb{R}^{N\times 1}$.
The $n$-th element of $\mathbf{y}$ is defined as ${y}_n=\boldsymbol{\mu}_k\mathbf{v}_D^\intercal ~  \text{if} ~  \mathbf{x}_n \in  \mathcal{C}_k$, i.e., the projection of $\mathbf{x}_n$'s cluster mean into the direction of the largest eigenvector.
The objective is to minimize the difference between spectral embeddings $f(\mathbf{X})$ and the target matrix $\mathbf{Y}$ in the direction of $\mathbf{v}_D$:
\begin{equation}
	\label{eqn:finetune}
	\mathcal{L}_{\text{greedy}} =\left\|f(\mathbf{X})\mathbf{v}_D^\intercal  - \mathbf{y} \right\|^2
	= \left\| (f(\mathbf{X})\mathbf{V} - \mathbf{Y} )\mathbf{I}_{(-1)} \right\|^2
\end{equation}
where $\mathbf{I}_{(-1)} $ is the last column of the identity matrix $\mathbf{I}$.
In this paper, we solve for $\mathbf{v}_D$ and minimize the Kmeans objective along $\mathbf{v}_D$ in an alternate manner, thus terming the above equation as greedy Kmeans loss.

\begin{table*}[t]
	\caption{Clustering performance (ACC\% and NMI\%) comparison.
		%		Dash mark (-) denotes unavailable results due to memory limitation.
		The best and runner-up results are highlighted in \textbf{bold} and \underline{underline}, respectively.}
	\centering
	%	\resizebox{\textwidth}{!}{
		\begin{tabular}{lm{6mm}m{6mm}m{6mm}m{6mm}m{6mm}m{6mm}m{6mm}m{6mm}m{6mm}m{6mm}m{6mm}m{6mm}m{6mm}m{6mm}m{6mm}}
			\toprule
			\multicolumn{1}{c}{Dataset} & \multicolumn{2}{c}{USPS} & \multicolumn{2}{c}{FASHION} & \multicolumn{2}{c}{FASHION-{test}} & \multicolumn{2}{c}{MNIST} & \multicolumn{2}{c}{MNIST-{test}} & \multicolumn{2}{c}{COIL20} &\multicolumn{2}{c}{FRGC}  \\
			\multicolumn{1}{c}{Method} & ACC & NMI & ACC & NMI & ACC & NMI & ACC & NMI & ACC & NMI & ACC & NMI& ACC & NMI\\
			\midrule
			Kmeans & 66.81 & 62.62 & 47.57 & 51.22 & 52.25 & 58.78 & 53.22 & 49.96 & 54.24 & 50.00 & 25.28 & 67.17&12.63&18.27  \\
			SC-Ncut & 62.51 & 69.48 & 52.06 & 59.05 & 52.24 & 58.76 & 63.15 & 73.25 & 59.54 & 69.12 & 60.35 & 72.42&27.50&33.03 \\
			SC-NJW & 66.91 & 61.66 & 55.17 & 62.82 & 46.15 & 50.31 & 55.17 & 62.82 & 56.13 & 48.62 & 70.14 & 79.42&42.45&51.52\\
			PIC & 61.79 & 58.67 & 44.18 & 54.21 & 46.74 & 50.59 & 49.89 & 60.67 & 51.23 & 54.05 & 18.61 & 38.52&16.04&11.44 \\
			FUSE & 57.55 & 56.86 & 52.10 & 53.44 & 49.00 & 52.00 & 63.89 & 69.64 & 77.00 & 72.00 & 23.47 & 37.17 &17.43&12.25\\
			SE-ISR & 60.55 & 65.60 & 57.38 & 57.22 & 51.44 & 55.15 & 57.56 & 55.06 & 67.36 & 64.25 & 69.51 & 82.79 &36.23&44.28\\
			%			\cmidrule(lr){1-15}
			DEC & 69.75 & 68.66 & 51.80 & 55.23 & {{59.20}} & 59.52 & 81.27 & 75.70 & 75.86 & 69.89 & 53.47 & 70.81&32.13&40.64\\
			DEKM & 78.87 & 80.51 & 
			{58.89} & {{62.55}} & 52.36 & 60.27 & 95.75 & 91.06 & 83.99 & 80.86 & 72.62 & 81.97&37.29&49.32\\
			Graphencoder & 20.66 & 12.65 & -& - & 34.28 & 31.19 & 14.84 & - & - & 5.55 & 45.49 & 59.87  &12.84&7.42 \\
			SpectralNet & 78.12 & 81.36 & 53.07 & 59.07 & 52.58 & 58.36 & 60.95 & 67.02 & 61.91 & 68.21 & 64.03
			& {86.39} &31.52&43.60\\
			DEPICT & \underline{{96.40}} & \underline{{92.70}} & 50.38 & 57.04 & 50.74 & 51.74 & {{96.50}} & {{91.70}} & \underline{{96.50}} & {{91.50}} & {74.38} & 81.97&\underline{{47.00}}&\textbf{{61.00}}\\
			SENet &  80.60 &  75.37 & \textbf{66.19}  & \underline{66.08} &\underline{60.24}& \underline{64.14} & \underline{96.80}  & 91.69  & 96.27  & \underline{91.94}   &  44.58 & 71.06  & {46.72}   & 27.71   \\
			EDESC & 72.90 & 69.20 & 55.13 & 58.41 & 56.91 & 56.79 & 74.15 & 66.74 & 66.23 & 56.41 & 70.42& 78.16 &28.76&30.53\\
			WEC-MMD & -& -&  62.20 & 62.96  & -& -& 96.74 & \underline{92.23} & -  &  -  & \underline{82.71} & \underline{87.80}  & -   &  -  \\
			DML-DSL & 84.41 & 79.47& 63.20 & 64.80   &55.34 &57.50 & 96.33 & 91.22     &  95.16& 89.23   & \textbf{88.38}  &  70.54 &  \textbf{79.12}  &  46.75  \\
			DSC & \textbf{{97.26}} & \textbf{{93.00}} & \underline{{64.62}} & \textbf{{67.98}} & \textbf{{62.56}} & \textbf{{68.81}} & \textbf{{97.80}} & \textbf{{94.10}} & \textbf{{96.86}} & \textbf{{92.36}} & {{81.88}} & \textbf{{89.53}}&41.55&\underline{{58.27}} \\
			\bottomrule
		\end{tabular}
		%	}
	\label{tab_all}
\end{table*}

\subsection{Joint Loss Function}\label{sec:step3}
We consider jointly optimizing the spectral embedding loss and greedy Kmeans loss in an end-to-end way.
To this end, we slightly modify the spectral embedding Loss \eqref{eqn:E_H}.
Instead of forcing the encoder to predict the whole spectral embeddings $\mathbf{Z}$, we encourage the encoder to predict $\mathbf{Z}$ in all directions except for the direction that contains the worst cluster structures:
\begin{equation}
	\label{eq_loss1}
	\mathcal{L}_{\text {spectral}}=\left\| \left(f(\mathbf{X})\mathbf{V} - \mathbf{Y}\right)\mathbf{I}_{(D-1)}  \right\|^2_F
\end{equation}
where we use the fact the first $D-1$ columns of $\mathbf{Y}$ and $\mathbf{Z}\mathbf{V}$ are the same, and $\mathbf{I}_{(D-1)}$ is the first $D-1$ columns of the identity matrix $\mathbf{I}$.
This above loss can still learn meaningful embeddings, as most discriminative information exists in the directions with good cluster structures.
Finally, we combine the above loss with the greedy Kmeans Loss \eqref{eqn:finetune} seamlessly:
\begin{equation}
	\label{eq_loss_all}
	\mathcal{L}_{\text {joint}}= \mathcal{L}_{\text {spectral}}+\mathcal{L}_{\text {greedy}}
	= \left\|f(\mathbf{X})\mathbf{V} - \mathbf{Y}  \right\|^2_F
\end{equation}

Algorithm \ref{algo} shows the pseudo-code of DSC.
The time complexity of DSC for training is $\mathcal{O}(NB+2N^2D+N^2DT+NDK+ND+D^3+ND)$ and for inference is $\mathcal{O}(NDK)$, where $B$ denotes the number of neurons in the autoencoder.
Since $B$, $D$, and $T$ are constants, the time complexity for training is bounded by $\mathcal{O}(N^2K)$ and for inference is bounded by $\mathcal{O}(NK)$.
In addition, the space complexity is $\mathcal{O}(N^2K)$.

\begin{table*}[thbp]
	\centering
	\caption{Ablation study of DSC. }
	%	\resizebox{\textwidth}{!}{
		\begin{tabular}{lm{6mm}m{6mm}m{6mm}m{6mm}m{6mm}m{6mm}m{6mm}m{6mm}m{6mm}m{6mm}m{6mm}m{6mm}m{6mm}m{6mm}m{6mm}}
			\toprule
			\multicolumn{1}{c}{Dataset} & \multicolumn{2}{c}{USPS} & \multicolumn{2}{c}{FASHION} & \multicolumn{2}{c}{FASHION-{test}} & \multicolumn{2}{c}{MNIST} & \multicolumn{2}{c}{MNIST-{test}} & \multicolumn{2}{c}{COIL20} &\multicolumn{2}{c}{FRGC}  \\
			\multicolumn{1}{c}{Method} & ACC & NMI & ACC & NMI & ACC & NMI & ACC & NMI & ACC & NMI & ACC & NMI& ACC & NMI\\
			\midrule
			AE+Kmeans & 83.93 & 80.05 & 56.36 & 61.87 & 60.40 & 62.30 & 94.43 & 87.51 & 91.07 & 81.87 & 68.72 & 80.34&37.04&46.15 \\
			AE+SC-Ncut & 76.44 & 74.86 & 56.33 & 59.38 & 54.76 & 59.56 & 93.59 & 86.29 & 83.75 & 76.61 & 69.91 & 80.51&41.52&52.29\\
			AE+SE & 96.21 & 91.29 & 64.23 & 67.63 & 62.50 & 68.65 & 97.34 & 93.28 & 96.01 & 90.92 & 77.02 & 86.97&40.68&56.84 \\
			AE+GK &  86.67 & 82.15 & 56.91 & 62.25 & 61.49 & 62.84 & 95.68 & 89.68 & 93.37 & 85.54 & 71.56 & 80.85 &41.17&54.06\\
			AE+SE+GK (DSC) & \textbf{97.26} & \textbf{93.00} & \textbf{64.62} & \textbf{67.98} & \textbf{62.56} & \textbf{68.81} & \textbf{97.80} & \textbf{94.10} & \textbf{96.86} & \textbf{92.36} & \textbf{81.88} & \textbf{89.53}&\textbf{41.55}&\textbf{58.27}  \\ 
			\bottomrule  
		\end{tabular}            
		%	}
	\label{tab_ablation}
\end{table*}

\section{Experimental Evaluation}
In this section, we first compare DSC with 15 clustering methods on 7 benchmark datasets. We then conduct ablation study to evaluate the effect of each module of DSC. Subsequently, we evaluate the generalization ability by conducting out-of-sample experiment. After that, we evaluate the efficiency of each method by comparing their running time. Finally, we visualize the learned embeddings at different training stages.

\noindent\textbf{Datasets and Metrics.}
We utilize 7 benchmark datasets that cover handwritten digits, fashion products, multi-view objects, and human faces:
USPS \cite{hull1994database}, MNIST\cite{lecun1998gradient}, FASHION \cite{xiao2017fashion}, COIL-20 \cite{nene1996columbia}, FRGC \cite{yang2016joint}.
The first three datasets each contains 10 clusters, while the latter two datasets each contains 20 clusters.
Unless explicitly specified, we concatenate the training and test sets of each dataset for evaluation.
We use two standard metrics to evaluate clustering performance: Clustering Accuracy (ACC) and Normalized Mutual Information (NMI).
Both metrics range from 0 to 1, and higher values indicate better clustering performance.

\noindent\textbf{Implementation Details.}
We employ a convolutional autoencoder for evalution.
The encoder consists of three stacked convolutional layers with 32, 64, and 128 channels respectively and a linear embedding layer with 10 neurons. 
The decoder is a mirrored version of the encoder layers.
The parameter $M$ for constructing the self-tuning affinity matrix $\sigma$ is set to 7.
The early stopping threshold $\hat{a}$ of power iteration is set to 0.01, and the maximum update iteration number is limited to 15.
The batch size is set to 256 for pretraining the autoencoder and 32 for clustering training. 
In each iteration of clustering training, we use 40 mini-batches of samples.
The autoencoder is trained by the Adam optimizer with parameters $lr=0.001$, ${\beta_1} = 0.9$, ${\beta_2} = 0.999$. 
We pretrain the autoencoder with 200 epochs and stop the clustering training when less than 0.5\% of samples change their cluster assignments between two consecutive iterations. 
We fix these parameter settings for all the datasets.
Our code is available at \url{https://github.com/spdj2271/DSC}.

\subsection{Comparisons to State-of-the-Art Methods}
% We compare with 6 traditional and 9 deep clustering methods: Kmeans~\cite{lloyd1982least}, SC-Ncut \cite{shi2000normalized}, SC-NJW \cite{ng2002spectral}, PIC \cite{lin2010power}, FUSE \cite{ye2016fuse}, SE-ISR \cite{wang2021fast}, DEC \cite{DEC_xie2016unsupervised}, DEKM \cite{guo2021deep}, Graphencoder \cite{tian2014learning}, SpectralNet \cite{shaham2018spectralnet}, DEPICT \cite{ghasedi2017deep}, SENet \cite{zhang2021learning}, EDESC \cite{cai2022efficient}, DML-DS \cite{sadeghi2023deep}, and WEC-MMD \cite{cai2024wasserstein}.
We compare the proposed DSC with 6 conventional clustering methods and 9 deep clustering methods. The conventional clustering methods contain: Kmeans~\cite{lloyd1982least}; two common spectral clustering methods: spectral clustering with normalized cuts (SC-Ncut) \cite{shi2000normalized} and NJW spectral clustering (SC-NJW) \cite{ng2002spectral}; three improved spectral clustering methods: power iteration clustering (PIC) \cite{lin2010power}, full spectral clustering (FUSE) \cite{ye2016fuse}, spectral clustering with simultaneous spectral embedding and spectral rotation (SE-ISR) \cite{wang2021fast}.
The deep clustering methods contain: deep embedded clustering (DEC) \cite{DEC_xie2016unsupervised}, 
deep embedded Kmeans clustering (DEKM) \cite{guo2021deep}, autoencoder-based spectral graph clustering (Graphencoder) \cite{tian2014learning}, spectral clustering network (SpectralNet) \cite{shaham2018spectralnet}, deep embedded regularized clustering (DEPICT) \cite{ghasedi2017deep}, self-expressive clustering network (SENet) \cite{zhang2021learning}, deep embedded subspace clustering (EDESC) \cite{cai2022efficient}, deep multi-representation clustering (DML-DS) \cite{sadeghi2023deep}, and deep Wasserstein embedding clustering (WEC-MMD) \cite{cai2024wasserstein}.

%As shown in Table \ref{tab_all}, DSC achieves competitive results on all these datasets.
Table \ref{tab_all} shows the overall clustering results on the 7 benchmark datasets.
DSC achieves much better clustering performance than all the shallow clustering methods (the first 6 rows in Table \ref{tab_all}). 
%Similar to DSC, SC-ISR optimizes spectral embedding and spectral rotation simultaneously, but it is significantly outperformed by DSC. This large margin in performance is due to the powerful representation capability of the neural networks.
Similar to DSC, SC-ISR optimizes spectral embedding and spectral rotation simultaneously, but DSC significantly outperforms it. 
This large margin in performance is due to the powerful representation capability of the neural networks.
Compared with deep clustering methods, DSC achieves better results on most datasets, which can be attributed to the effective simultaneous representation (spectral embedding) learning and clustering strategy.
%Thus, DSC is a more promising clustering method compared with the alternatives in practical applications.

\subsection{Ablation Study}
DSC consists of two main modules: the spectral embedding (SE) module and the greedy Kmeans (GK) module. 
To evaluate each module's effectiveness, we first detect clusters on the pretrained autoencoder embeddings with Kmeans (AE+Kmeans) and SC-Ncut (AE+SC-Ncut), whose clustering results serve as the baselines. 
We then train a model (AE+SE) by only minimizing $\mathcal{L}_{\text {spectral}}$ and report clustering results by conducting posthoc Kmeans clustering on the learned spectral embeddings.
Subsequently, we train another model (AE+EM) by only minimizing $\mathcal{L}_{\text {greedy}}$, where Kmeans is applied on the autoencoder embeddings to obtain the rotation matrix and generate the target matrix.
Table \ref{tab_ablation} indicates that both the SE and GK modules improve the clustering performance. 
The SE module plays a key role, which helps to reduce the dimensionality of the data while capturing the essential cluster structural information.
The GK module could improve the performance to some extent, which aligns with its purpose of fine-tuning the spectral embeddings to be Kmeans-friendly.

\begin{table}[t]
	\centering
	\caption{Clustering performance on out-of-sample data.}
	\resizebox{1.05\columnwidth}{!}{
		\begin{tabular}{m{10mm}m{13mm}m{13mm}m{13mm}m{13mm}m{1mm}}
			\toprule
			Dataset & \multicolumn{2}{c}{MNIST-test$\rightarrow$MNIST-train} & \multicolumn{3}{c}{FASHION-test$\rightarrow$FASHION-train} \\
			Method& \multicolumn{1}{c}{ACC} & \multicolumn{1}{c}{NMI} & \multicolumn{1}{c}{ACC} & \multicolumn{1}{c}{NMI} \\
			\midrule
			SpectralNet & 61.91$\rightarrow$61.31 & 68.21$\rightarrow$66.35 & 52.58$\rightarrow$52.74 & 58.36$\rightarrow$58.88 \\
			DSC & \textbf{96.86$\rightarrow$96.06} & \textbf{92.36$\rightarrow$90.59} & \textbf{62.56$\rightarrow$62.62} & \textbf{69.30$\rightarrow$68.78}\\
			\bottomrule
		\end{tabular}
	}
	\label{tab_oose}
\end{table}

\begin{figure}[t] 
	\centering
 \centerline{\includegraphics[width=\columnwidth]{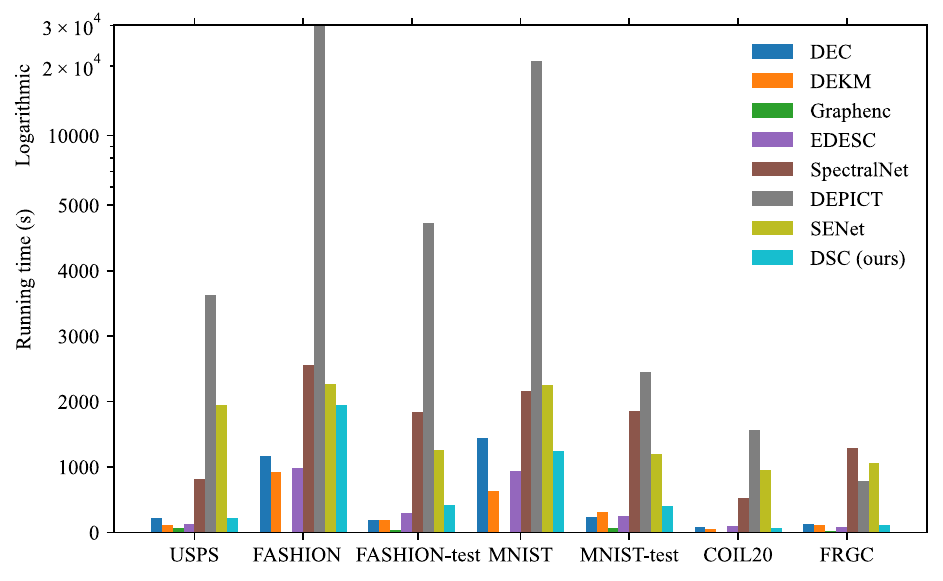}}
	\caption{Running time comparison (in seconds).}
	\label{fig_runningtime}
\end{figure}

\subsection{Evaluation of Generalization Ability} 
\label{sec_generalization}
We compare the generalization ability of our DSC with  the existing deep spectral clustering method  SpectralNet  by applying the trained models to the previously-unseen samples.
We respectively train DSC and SpectralNet on FASHION-{{test}} and MNIST-{{test}} datasets and then evaluate these two models on the FASHION-{train} and MNIST-{train} datasets. 
For a more challenging evaluation setting, we select to conduct the generalization experiments on the larger train datasets (60,000 samples) rather than the test datasets (10,000 samples).
Table~\ref{tab_oose} shows that DSC exhibits better generalization performance compared to SpectralNet.
%	This is because DSC uses the encoder to learn the spectral embeddings of the {test} dataset, and once trained the encoder can be directly applied to the {train} dataset without needing to compute and eigendecompose its graph Laplacian matrix from scratch.

\begin{figure}[t]
	\centering
	\hspace*{\fill}
	\subfigure[Raw samples]{\includegraphics[width=0.315\columnwidth]{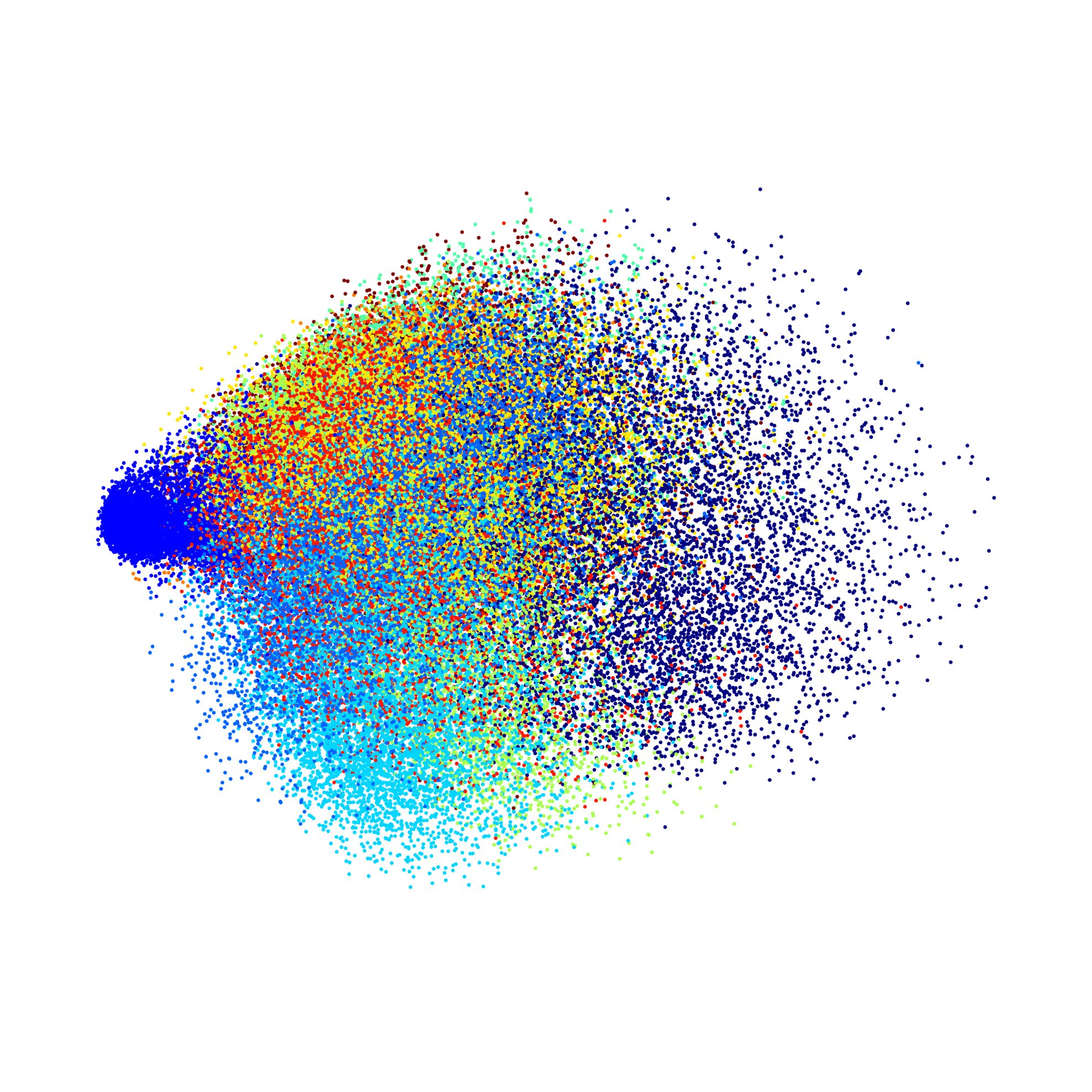}\label{fig_embed_a}}
	\hfill
	\hfill
	\subfigure[Autoencoder]{\includegraphics[width=0.315\columnwidth]{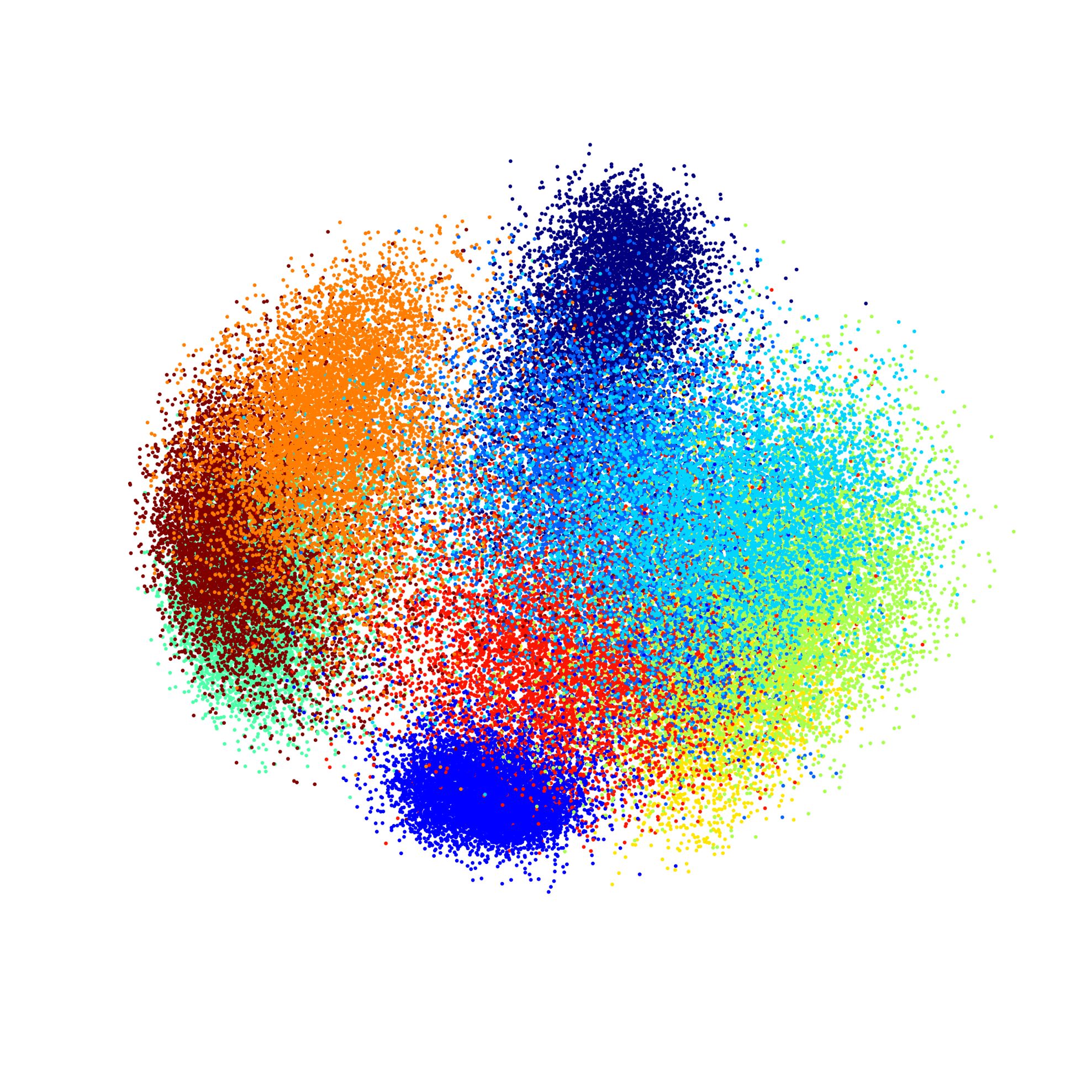}\label{fig_embed_b}}
	\hfill
	\hfill
	\subfigure[DSC]{\includegraphics[width=0.315\columnwidth]{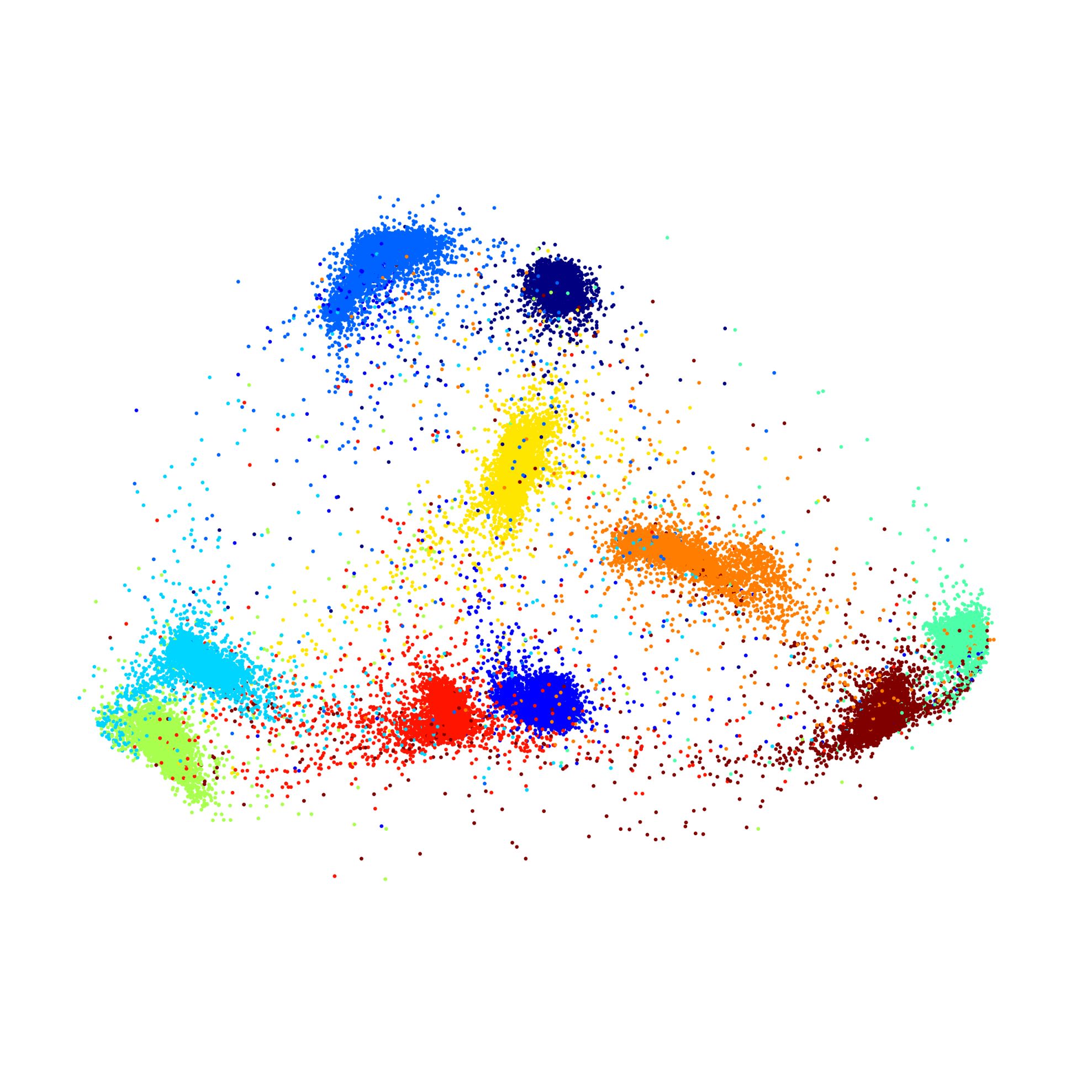}\label{fig_embed_c}}
	\hspace*{\fill}
	\caption{Embedding visualization using PCA.
	}
	\label{fig_embed}
\end{figure}

\subsection{Running Time Comparison}
% We compare the running time (including the time for pretraining the autoencoder) of our DSC with SpectralNet.
% The experiment is conducted on a server equipped with one Nvidia GeForce GTX 1050Ti GPU.
% As shown in Fig. \ref{fig_runningtime}, our DSC outperforms SpectralNet with a clear advantage, which is due to the use of the lightweight power iteration method in DSC.
% The majority of the running time for DSC is spent on pretraining the autoencoder, which is essential for any autoencoder-based clustering method.
We compare the running time (including the pretraining time of autoencoders) of each deep clustering model on all the datasets and the results are presented in Figure \ref{fig_runningtime}.
The experiment is conducted on a server equipped with one Nvidia GeForce GTX 1050Ti GPU.
Our DSC outperforms SpectralNet, DEPICT, and SENet with an obvious advantage.
Specifically, the running time of DEPICT dramatically grows on large datasets (FASHION and MNIST), leading to manual stopping after 30,000 seconds on the FASHION dataset.
In contrast, DSC only takes 1,900 seconds to complete the FASHION dataset. 
Despite the efficiency advantages exhibited by other competitors, such as DEC, DEKM, Graphenc, and EDESC, their clustering performance is clearly unsatisfactory as demonstrated in Table \ref{tab_all}.
Our DSC requires a similar running time to these models but achieves significantly higher clustering performance. In fact, the majority running time of DSC expends on the pretraining of autoencoder, which is requisite for any autoencoder-based clustering methods.

\subsection{Embedding Space Comparison}
We visualize the embedding space at different training stages of DSC on the MNIST dataset in Fig. \ref{fig_embed}.
We conduct Principal Component Analysis (PCA) \cite{pearson1901liii} on the embeddings and then select the first three principal components for visualization. 
Fig. \ref{fig_embed_a} shows that raw samples are in a chaotic status and there are no obvious cluster structures. 
Compared with the pretrained autoencoder embeddings in Fig. \ref{fig_embed_b}, the embeddings learned by DSC in Fig. \ref{fig_embed_c} are well-separated, exhibiting much better cluster structures.

	\section{Conclusion}
In this paper, we have proposed a novel deep spectral clustering method called DSC. It combines the deep autoencoder and power iteration to efficiently learn the discriminative spectral embeddings.
DSC also introduces a greedy Kmeans objective to make the spectral embeddings Kmeans-friendly.
Experiments on seven real-world datasets indicate that DSC achieves state-of-the-art clustering performance.
In the future, we will extend DSC to cluster multi-view data.

\section{Acknowledgments}
We thank the anonymous reviewers for their valuable and constructive comments. 
This work was supported partially by the National Natural Science Foundation of China (grant \#62176184), the National Key Research and Development Program of China (grant \#2020AAA0108100), and the Fundamental Research Funds for the Central Universities of China.

	\bibliographystyle{ieeetr}
	\bibliography{reference}

\begin{thebibliography}{10}

\bibitem{lloyd1982least}
S.~Lloyd, ``Least squares quantization in pcm,'' {\em IEEE Trans. Inf. Theory},
  1982.

\bibitem{xiao2017fashion}
H.~Xiao, K.~Rasul, and R.~Vollgraf, ``Fashion-mnist: a novel image dataset for
  benchmarking machine learning algorithms,'' 2017.

\bibitem{wang2015spectral}
S.~Wang, F.~Chen, and J.~Fang, ``Spectral clustering of high-dimensional data
  via nonnegative matrix factorization,'' in {\em IJCNN}, 2015.

\bibitem{pearson1901liii}
K.~Pearson, ``Liii. on lines and planes of closest fit to systems of points in
  space,'' {\em The London, Edinburgh, and Dublin Philosophical Magazine and
  Journal of Science}, 1901.

\bibitem{lee1999learning}
D.~D. Lee and H.~S. Seung, ``Learning the parts of objects by non-negative
  matrix factorization,'' {\em Nature}, 1999.

\bibitem{yang2016unified}
Y.~Yang, F.~Shen, Z.~Huang, and H.~T. Shen, ``A unified framework for discrete
  spectral clustering.,'' in {\em IJCAI}, 2016.

\bibitem{pang2018spectral}
Y.~Pang, J.~Xie, F.~Nie, and X.~Li, ``Spectral clustering by joint spectral
  embedding and spectral rotation,'' {\em TCYB}, 2018.

\bibitem{tian2014learning}
F.~Tian, B.~Gao, Q.~Cui, E.~Chen, and T.-Y. Liu, ``Learning deep
  representations for graph clustering,'' in {\em AAAI}, 2014.

\bibitem{shaham2018spectralnet}
U.~Shaham, K.~Stanton, H.~Li, R.~Basri, B.~Nadler, and Y.~Kluger,
  ``Spectralnet: Spectral clustering using deep neural networks,'' in {\em
  ICLR}, 2018.

\bibitem{shi2000normalized}
J.~Shi and J.~Malik, ``Normalized cuts and image segmentation,'' {\em TPAMI},
  2000.

\bibitem{ng2002spectral}
A.~Y. Ng, M.~I. Jordan, Y.~Weiss, {\em et~al.}, ``On spectral clustering:
  Analysis and an algorithm,'' {\em NeurIPS}, 2002.

\bibitem{lin2010power}
F.~Lin and W.~W. Cohen, ``Power iteration clustering,'' in {\em ICML}, 2010.

\bibitem{boutsidis2015spectral}
C.~Boutsidis, P.~Kambadur, and A.~Gittens, ``Spectral clustering via the power
  method-provably,'' in {\em ICML}, 2015.

\bibitem{huang2015diverse}
H.~Huang, S.~Yoo, D.~Yu, and H.~Qin, ``Diverse power iteration embeddings:
  Theory and practice,'' {\em TKDE}, 2015.

\bibitem{ye2016fuse}
W.~Ye, S.~Goebl, C.~Plant, and C.~B{\"o}hm, ``Fuse: Full spectral clustering,''
  in {\em SIGKDD}, 2016.

\bibitem{learned2003ica}
E.~G. Learned-Miller and J.~W.~F. Iii, ``Ica using spacings estimates of
  entropy,'' {\em J. Mach. Learn. Res.}, 2003.

\bibitem{householder1958unitary}
A.~S. Householder, ``Unitary triangularization of a nonsymmetric matrix,'' {\em
  Journal of the ACM (JACM)}.

\bibitem{DEC_xie2016unsupervised}
J.~Xie, R.~Girshick, and A.~Farhadi, ``Unsupervised deep embedding for
  clustering analysis,'' in {\em ICML}, 2016.

\bibitem{yang2016joint}
J.~Yang, D.~Parikh, and D.~Batra, ``Joint unsupervised learning of deep
  representations and image clusters,'' in {\em CVPR}, 2016.

\bibitem{ghasedi2017deep}
K.~Ghasedi~Dizaji, A.~Herandi, C.~Deng, W.~Cai, and H.~Huang, ``Deep clustering
  via joint convolutional autoencoder embedding and relative entropy
  minimization,'' in {\em ICCV}, 2017.

\bibitem{zhang2021learning}
S.~Zhang, C.~You, R.~Vidal, and C.-G. Li, ``Learning a self-expressive network
  for subspace clustering,'' in {\em CVPR}, 2021.

\bibitem{cai2022efficient}
J.~Cai, J.~Fan, W.~Guo, S.~Wang, Y.~Zhang, and Z.~Zhang, ``Efficient deep
  embedded subspace clustering,'' in {\em CVPR}, 2022.

\bibitem{kwon2023image}
S.~Kwon, J.~Park, M.~Kim, J.~Cho, E.~K. Ryu, and K.~Lee, ``Image clustering
  conditioned on text criteria,'' in {\em ICLR}, 2024.

\bibitem{zhang2023clusterllm}
Y.~Zhang, Z.~Wang, and J.~Shang, ``Clusterllm: Large language models as a guide
  for text clustering,'' in {\em EMNLP}, 2023.

\bibitem{miklautz2022deep}
L.~Miklautz, M.~Teuffenbach, P.~Weber, R.~Perjuci, W.~Durani, C.~B{\"o}hm, and
  C.~Plant, ``Deep clustering with consensus representations,'' in {\em ICDM},
  2022.

\bibitem{leiber2022dipencoder}
C.~Leiber, L.~G. Bauer, M.~Neumayr, C.~Plant, and C.~B{\"o}hm, ``The
  dipencoder: Enforcing multimodality in autoencoders,'' in {\em SIGKDD}, 2022.

\bibitem{leiber2021dip}
C.~Leiber, L.~G. Bauer, B.~Schelling, C.~B{\"o}hm, and C.~Plant, ``Dip-based
  deep embedded clustering with k-estimation,'' in {\em SIGKDD}, 2021.

\bibitem{miklautz2021details}
L.~Miklautz, L.~G. Bauer, D.~Mautz, S.~Tschiatschek, C.~B{\"o}hm, and C.~Plant,
  ``Details (don't) matter: Isolating cluster information in deep embedded
  spaces.,'' in {\em IJCAI}, 2021.

\bibitem{achiam2023gpt}
J.~Achiam, S.~Adler, S.~Agarwal, L.~Ahmad, I.~Akkaya, F.~L. Aleman, D.~Almeida,
  J.~Altenschmidt, S.~Altman, S.~Anadkat, {\em et~al.}, ``Gpt-4 technical
  report,'' 2023.

\bibitem{zhou2022comprehensive}
S.~Zhou, H.~Xu, Z.~Zheng, J.~Chen, J.~Bu, J.~Wu, X.~Wang, W.~Zhu, M.~Ester,
  {\em et~al.}, ``A comprehensive survey on deep clustering: Taxonomy,
  challenges, and future directions,'' 2022.

\bibitem{zelnik2004self}
L.~Zelnik-Manor and P.~Perona, ``Self-tuning spectral clustering,'' {\em
  NeurIPS}, 2004.

\bibitem{hull1994database}
J.~J. Hull, ``A database for handwritten text recognition research,'' {\em
  TPAMI}, 1994.

\bibitem{lecun1998gradient}
Y.~LeCun, L.~Bottou, Y.~Bengio, and P.~Haffner, ``Gradient-based learning
  applied to document recognition,'' 1998.

\bibitem{nene1996columbia}
S.~A. Nene, S.~K. Nayar, H.~Murase, {\em et~al.}, ``Columbia object image
  library (coil-100),'' 1996.

\bibitem{wang2021fast}
Z.~Wang, X.~Dai, P.~Zhu, R.~Wang, X.~Li, and F.~Nie, ``Fast optimization of
  spectral embedding and improved spectral rotation,'' {\em TKDE}, 2021.

\bibitem{guo2021deep}
W.~Guo, K.~Lin, and W.~Ye, ``Deep embedded k-means clustering,'' in {\em
  ICDMW}, 2021.

\bibitem{sadeghi2023deep}
M.~Sadeghi and N.~Armanfard, ``Deep multirepresentation learning for data
  clustering,'' {\em TNNLS}, 2023.

\bibitem{cai2024wasserstein}
J.~Cai, Y.~Zhang, S.~Wang, J.~Fan, and W.~Guo, ``Wasserstein embedding learning
  for deep clustering: A generative approach,'' {\em TMM}, 2024.

\end{thebibliography}

\end{document}